\documentclass[12pt]{article}

\usepackage{sbc-template}

\usepackage{cite}
\usepackage[table]{xcolor}
\usepackage[utf8x]{inputenc}
\usepackage{array} 
\usepackage{multirow}
\usepackage{amsmath}
\usepackage{xcolor}
\usepackage{subfigure}  
\usepackage{amsthm, amsfonts, amsmath, amssymb}
\usepackage{url}
\usepackage{graphicx}
\usepackage{booktabs}

\newcommand{\aspas}[1]{``#1''}
\newcommand{\til}{$\sim$}

\title{Evaluating Word Embeddings for Sentence Boundary Detection in Speech Transcripts}

\author{Marcos V. Treviso\inst{1}, Christopher D. Shulby\inst{1,2}, Sandra M. Aluísio\inst{1}}
\address{
Institute of Mathematics and Computer Science, University of S\~ao Paulo (USP) 
\nextinstitute
  CPqD
\email{marcostreviso@usp.br \ \{cshulby,sandra\}@icmc.usp.br}}

\begin{document}
\maketitle

\begin{abstract} 

This paper is motivated by the automation of neuropsychological tests involving discourse analysis in the retellings of narratives by patients with potential cognitive impairment. In this scenario the task of sentence boundary detection in speech transcripts is important as discourse analysis involves the application of Natural Language Processing tools, such as taggers and parsers, which depend on the sentence as a processing unit. Our aim in this paper is to verify which embedding induction method works best for the sentence boundary detection task, specifically whether it be those which were proposed to capture semantic, syntactic or morphological similarities.

\end{abstract}

\section{Introduction}

The concept of a sentence in written or spoken texts is important in several Natural Language Processing (NLP) tasks, such as morpho-syntactic analysis ~\cite {kepler2010, fonseca2016improving}, sentiment analysis~\cite{anchieta2015using, brum2016}, summarization~\cite{nobrega2016}, and speech processing~\cite{mendoncca2014method}, among others. However, punctuation marks that constitute a sentence boundary are ambiguous
The Disambiguation of Punctuation Marks (DPM) task analyzes punctuation marks in texts and indicates whether they correspond to a sentence boundary. The purpose of the DPM task is to answer the question: \textit{Among the tokens of punctuation marks in a text, which of them correspond to sentence boundaries?}

The Sentence Boundary Detection (SBD) task is very similar to DPM, both of which attempt to break a text into sequential units that correspond to sentences, where DPM is text-based and SBD can be applied to either written text or audio transcriptions and often for clauses, which do not necessarily end in final punctuation marks but are complete thoughts nonetheless. However, performing SBD in speech texts is more complicated due to the lack of information such as punctuation and capitalization; moreover text output is susceptible to recognition errors, in case of Automatic Speech Recognition (ASR) systems are used for automatic transcriptions~\cite{Gotoh2000}.
SBD from speech transcriptions is a task which has gained more attention in the last decades due to the increasing popularity of ASR software which automatically generate text from audio input. 
This task can also be applied to written texts, like online product reviews~\cite{Silla2004,Read2012,Lopez2015}, in order to better their intelligibility and facilitate the posterior use of NLP tools. 

It is important to point out that the differences between spoken and written texts are notable, mainly when we take into consideration the size of the utterances and the number of disfluencies provided in speech. Disfluencies include filled pauses, repetitions, modifications, repairs, partial utterances, nonword vocalizations and false starts. These phenomena are very common in spontaneous speech.\cite{Liu2004}.

\begin{figure}[!htb]
  \label{fig:exemplo}
  \centering
  \includegraphics[scale=0.45]{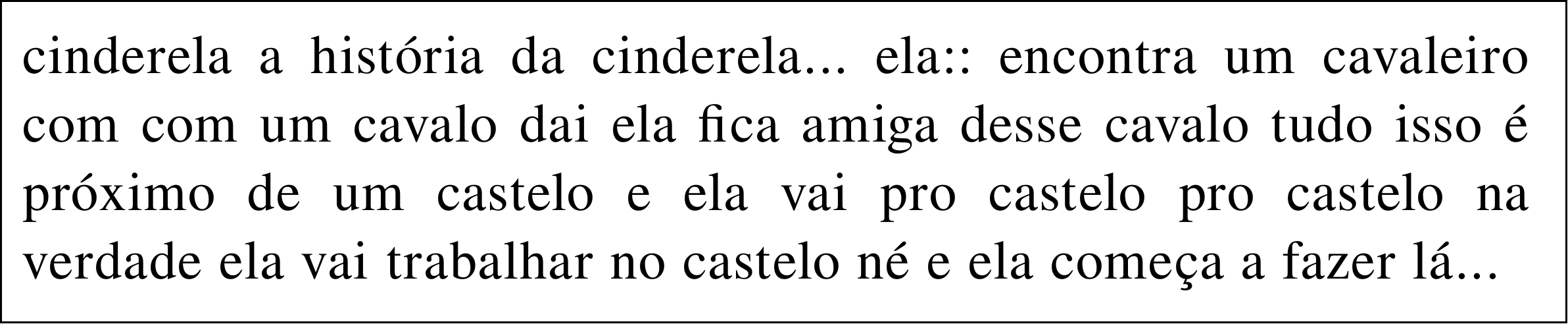}
   \caption{Narrative excerpt transcribed using the NURC annotation manual\protect\footnotemark{}}
\end{figure}

Figure \ref{fig:exemplo} shows the result of a transcript from a neuropsychological retelling task that does not include either capitalization or sentence segmentation, preventing the direct
application of NLP methods that rely on these marks for their correct use, such as taggers and parsers. One can easily note that this type of text differs greatly in style and form from written/edited text (on which most NLP tools are trained), such as text found in novels or a newspaper.

\footnotetext{\url{http://www.letras.ufrj.br/nurc-rj/}}

These tests are applied by clinicians who tell a story to patients who are instructed to try and remember as many details as possible so that they may retell it. The evaluation of language in discourse production, mainly in narratives, is an attractive alternative because it allows the analysis of linguistic microstructures and phonetic-phonological, morpho-syntactic, semantic-lexical components, as well as semantic-pragmatic macrostructures. 
Neuropsychological tests are used in clinical settings for detection, progress monitoring and treatment observation in patients with dementias. In an ideal scenario we would like to automate the application of neuropsychological tests and the discourse analysis of the retellings. 

NLP applications generally receive text as input; therefore, words can be considered the basic processing unit. In this case, it is important that they are represented in a way which carries the load of all relevant information. In the current approach used here, words are induced representations in a dense vector space. These representations are known as word embeddings; able to capture semantic, syntactic and morphological information from large unannotated corpora \cite{Mikolov2013b,Wang2015,Lai2015}. 
Various studies show that textual information is important for SBD \cite{Gotoh2000,Batista2012,Che2016}. Even though textual information is a strong indicator for sentence delimitation, boundaries are often associated with prosodic information \cite{Shriberg2000,Batista2012}, like pause duration, change in pitch and change in energy. However, the extraction of this type of information requires the use of high quality resources, and consequently, few resources with prosodic information are available. On the other hand, textual information can easily be extracted in large scale from the web. Textual information can also be represented in various ways for SBD, for example, n-gram based techniques have presented good results for SBD \cite{Gotoh2000,Kim2003,Favre2008}; however, in contrast to word embeddings, they are induced representations in a sparse vector space. 

Our aim in this paper is to verify which embedding induction method works best for the SBD task, specifically whether it be those which were proposed to capture semantic, syntactic or morphological similarities. For example, we imagine that methods that capture morphological similarities may benefit the SBD performance for impaired speech, since a large number of words produced in this type of speech are out-of-vocabulary words.The paper is organized as follows. Section 2 presents related work on SBD using word embeddings; Section 3 describes the word embedding models evaluated in this paper; Section 4 presents our experimental setup, describing the datasets, method, and preprocessing steps used; Section 5 presents our findings and discussions. Finally, Section 6 concludes the paper and outlines some future work.

\section{Related Work}

The work of \cite{Che2016} and \cite{TilkA15} use word embeddings to detect boundaries in prepared speech sentences, more specifically in the corpus from 2012 TED talks\footnote{\url{https://www.ted.com/talks}}. \cite{Che2016} propose a CNN (Convolution Neural Network)-based method with 50 dimensions using GloVe \cite{Pennington2014}. In \cite{Klejch2016,Klejch2017} the authors show that that textual information influences the retrieval of punctuation marks more than prosodic information, even without the use of word embeddings.

The work in \cite{TilkA15} is expanded in \cite{Tilk2016}, using bidirectional neural networks with attention mechanisms to evaluate a spontaneous telephone conversation corpus. The authors point out that the bidirectional vision of the RNN (Recurrent Neural Network) is a more impacting feature than the attention mechanism for SBD; with only the use of word embeddings, the achieved results yielded only 10\% less than when prosodic information was used together. In \cite{hough2017joint} a system that uses RNNs with word embeddings is proposed for the SBD task in conjunction disfluencies, where results are competitive with the state of the art are achieved on the Switchboard corpus \cite{Godfrey1992}, showing that the simultaneous execution of these tasks is superior to when done individually.  

Recently, the work of \cite{treviso2017sentence} proposed an automatic SBD method for impaired speech in Brazilian Portuguese, to allow a neuropsychological evaluation based on discourse analysis. The method uses RCNNs (Recurrent Convolutional Neural Networks) which independently treat prosodic and textual information, reaching state-of-the-art results for impaired speech. Also, this study showed that it is possible to achieve good results when comparing them with prepared speech, even when practically the same quantity of text is used. Another interesting evidence was that the use of word embeddings, without morpho-syntactic labels was able to present the same results as when they were used; this indicates that word embeddings contain sufficient morpho-syntactic information for SBD. It was also shown that the method gains the better results than the state-of-the-art method used by \cite{Fraser2015} by a great margin for both impaired and prepared speech (an absolute difference of \til0.20 and \til0.30, respectively). Beyond these findings, the method showed that the performance remains the same when a different story is used.

\section{Word Embeddings Models}

The generation of vector representations of words (or word embeddings) is linked to the induction method utilized. The work of \cite{Turian2010} divides these representations into three categories: cluster-based, distributional and distributed methods. In this paper, we focus only on distributed representations, because generally they are computationally faster to be induced. These representations are based on real vectors distributed in a multidimensional space induced by unsupervised learning. In the following paragraphs, we describe the three induction methods for word embeddings utilized in our evaluations.

A well-used NLP technique, Word2vec \cite{Mikolov2013b} follows the same principle as the natural language model presented in \cite{Collobert2008}, with the exception that it does not use a hidden layer, generating a computationally faster log-linear model. This technique is divided into two modeling types: (i) Continuous Bag-of-Words (CBOW), which given a window of words as input, the network tries to predict the word in the middle as output and (ii) the Skip-gram model, which tries to predict the window given the center word as input.

As Word2vec does not consider the word order in the window, this make the process less syntactic in nature, since word order is an essential phenomenon for syntax. In order to deal with this, a modification of Word2vec was proposed which is able to deal with word order by concatenating inputs in the CBOW model (instead of using the sum) and incremental weighting for Skip-gram. This technique is known as Wang2vec \cite{Wang2015}.

A recent induction technique called FastText \cite{Bojanowski2016,Joulin2016} uses n-grams of characters of a given word in the hope of capturing morphological information. In order to do this, the Skip-gram Word2vec model was modified so that that the scoring function of the network's output is calculated basing itself on the character n-gram vectors, which are summed with the context vectors in order to represent a word.

\section{Experimental Setup}

\subsection{Corpora/Datasets}


The datasets were divided into two categories: impaired speech and prepared speech. Impaired speech is not only spontaneous, but also noisy. The noise is produced internally due to the impaired neuropsychological condition of the participants studied. When people participate in neuropsychological tests, they produce the following phenomena: syntactically malformed sentences; mispronounced words (modifying the original morphology); low quality prosody (due to the shallow voices of the participants and/or abnormal fluctuations in vocal quality); and in general a great quantity and variety of types of disfluencies. 

The first dataset of discourse tests is a set of impaired speech narratives, based on a book of sequenced pictures from the well-known Cinderella story. 
This dataset consists of 60 narrative texts told by Brazilian Portuguese speakers; 20 healthy subjects, called controls (CTL), 20 patients with Alzheimer's disease (AD), and 20 patients with Mild Cognitive Impairment (MCI), diagnosed at Medical School of the University of São Paulo (FMUSP) and also used in \cite{Aluisio2016}. 
The second dataset was made available by the FalaBrasil project, and its contents are structured in the same way as the Brazilian Constitution from 1988 \cite{Batista2013constituicao}. The speech in this corpus can be categorized as prepared and also as read speech. 
To use these files in our scenario a preprocessing step was necessary, which removed lexical tips which indicate the beginning of articles, sections and paragraphs. This removal was carried out on both the transcripts and the audio. In addition, we separated the new dataset organized by articles, yielding 357 texts in total. 
Both datasets' properties are presented in Table \ref{tab:sumario_corpus}.

\begin{table}[!htb]
\centering
\footnotesize
\begin{tabular}{lll}
\toprule 
\textbf{Property}
& \textbf{Cinderela} 
& \textbf{Constitution} \\
\midrule
\# Transcipts 			& 60 		& 357   	\\
\# Words					& 23807 	& 63275 	\\
\# Sentences				& 2066 		& 2698   	\\
Duration						& 4h 11m	& 7h 39m 	\\
\bottomrule
\end{tabular}
\caption{Summary of corpora utilized in the experiments.}
\label{tab:sumario_corpus}
\end{table}	


The corpus used to induce the vectors is made up of text from Wikipedia in 2016, a news crawler which collected articles from the G1 portal\footnote{\url{http://g1.globo.com/}} and the PLN-BR \cite{bruckschen2008anotaccao} corpus. We also executed some basic preprocessing steps on this corpus, being that we forced all of the text to lowercase forms and separated each token from punctuation marks and tokenized the text using whitespace. We do not remove stopwords. After these steps, the embedding induction on the corpus returned \til356M tokens, of which \til1.7M were distinct.




\subsection{Method}



\begin{figure*}[!htb]
  \centering
  \includegraphics[width=\textwidth]{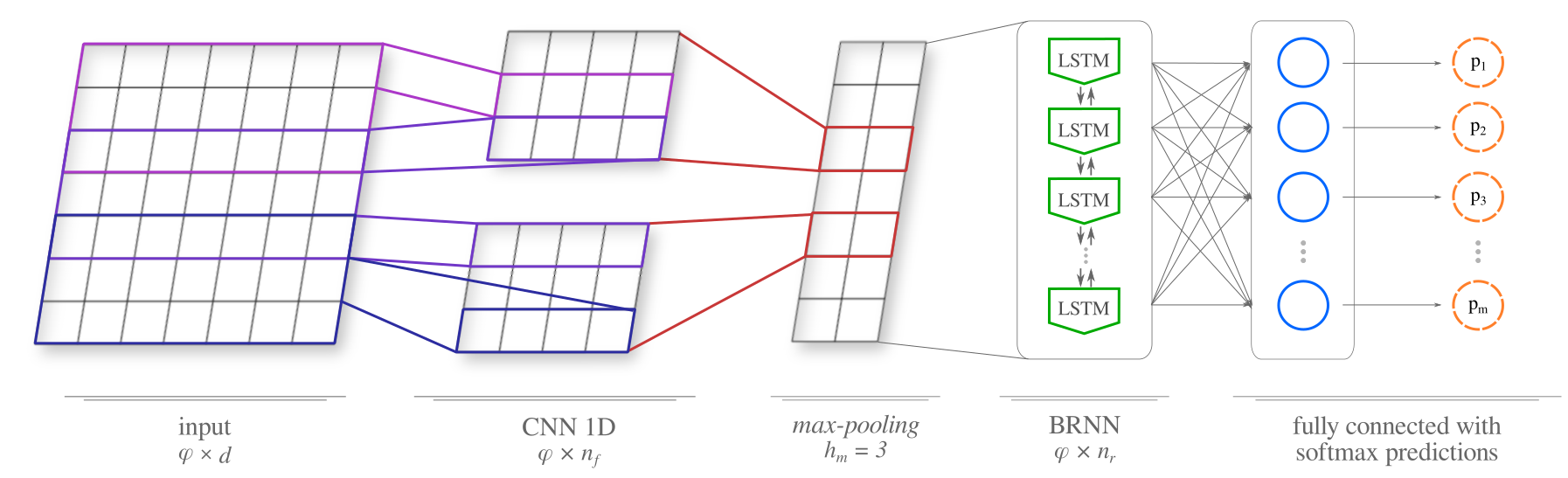}
  \caption{Architecture adapted from \cite{treviso2017sentence}}
  \label{fig:rcnn}
\end{figure*}

In order to automatically extract new features from the input and at the same time deal with the long dependency problems between words, we propose a method based on RCNNs which was inspired by the sentence segmentation work done by \cite{TilkA15} and \cite{Che2016}, and also by the work on text classification utilizing RCNNs by \cite{Lai2015}, where we made some adaptations so that the basic unit of classification was a data sequence. The architecture of our RCNN is the same as the one used in \cite{treviso2017sentence} and can be seen in Figure \ref{fig:rcnn}.

The final model in \cite{treviso2017sentence} consists of a linear combination between a model which deals only with lexical information and another which treats only prosodic information. In this paper, we ignore the prosodic model and focus only on the textual information provided by the word embeddings. The strategy to utilize only this information is based on the idea that one can train a text-based model with a large amount of data, since text is readily found on the web.



The model's input is a tridimensional word embedding matrix $\mathbf{E} \in \mathbb{R}^{m \times \varphi \times d}$, where $m$ is equal to the vocabulary size used for training the embeddings.  
Once we have an input matrix composed by word embeddings, the convolutional layer extracts $n_f$ new features from a sliding window with the size $h_c$, which corresponds to the size of the filter applied to the concatenated vectors $[e_1,...,e_{h_c}]$ corresponding to a region of $h_c$ neighboring embeddings \cite{kim2014convolutional}. 

The convolutional layer produces features for each $t$-th word as it applies the shared filter for a window of $h_c$ embeddings $\mathbf{e}_{t-h_c+1:t}$ in a sentence with the size $\varphi$. Our convolutional layer moves in a single vertical dimension (CNN 1D), one step at a time, which results in a quantity of filters $q_f$ equal to $\varphi - h_c + 2*p + 1$. And since we want to classify exactly $\varphi$ elements, we added $p = \lfloor h_c / 2 \rfloor$ elements of padding to both sides of the sentence. In addition, we applied a max-pooling operation on the temporal axis focusing on a region of $h_m$ words, with the idea of feeding only the most important features to the next layer.

The features selected by the max-pooling layer are fed to a recurrent layer. The values of the hidden units are computed utilizing $n_r$ LSTM cells \cite{Hochreiter1997} defined as activation units. As in \cite{Tilk2016}, our recurrent layer is based on anterior and posterior temporal states using the bidirectional recurrent mechanism (BRNN). With the use of a bidirectional layer which treats convolutionized features, the network is adept at exploring the principal that nearby words usually have a great influence, while considering that distant words, either to the left or right, can also have an impact on the classification. This frequently happens in the SBD task, for example, in the case of interrogatives, question words like \aspas{quem} (\aspas{\textit{who}}), \aspas{qual} (\aspas{\textit{what}}) and \aspas{quando} (\aspas{\textit{when}}) can define a sentence.

After the BRNN layer, we use dropout as a regularization strategy, which attempts to prevent co-adaptation between hidden units during forward and back-propagation, where some neurons are ignored with the purpose of reducing the chance of overfitting \cite{Srivastava2014}. 
Finally, the last layer, receives the output of the BRNN for each instance $t$, and feeds each into a simple fully connected layer which produces predictions using the softmax activation function, which gives us the final probability that a word precedes a sentence boundary ($B$) or not ($NB$).

The word embeddings matrix $\mathbf{E}$ was adjusted during training. Our RCNN uses the same hyperparameters described in \cite{treviso2017sentence} and the same training strategy, which consists of cost-function minimization utilizing the RMSProp procedure \cite{Tieleman2012} with back-propagation, considering the unbalanced task of sentence segmentation by penalizing errors from the minority class harsher ($B$).

\section{Results and Discussion}

We ran a 5-fold cross-validation for each group analyzed (CLT, MCI or AD), which left about 10\% of the data for testing, the rest for training. 

\begin{figure}[!htb]
  \centering
  \subfigure[CTL]{\includegraphics[width=0.485\textwidth, keepaspectratio]{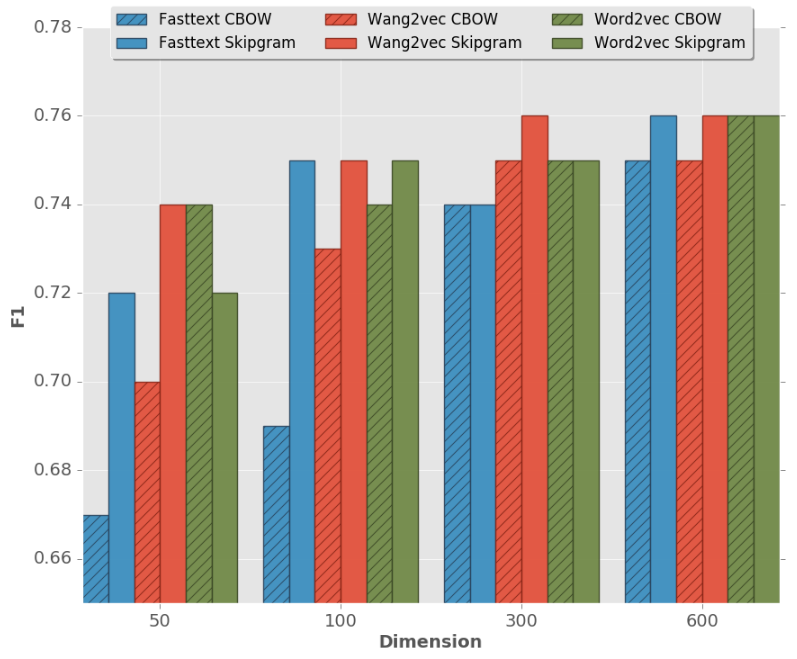} \label{fig:embs_ctl}} 
  \subfigure[MCI]{\includegraphics[width=0.485\textwidth, keepaspectratio]{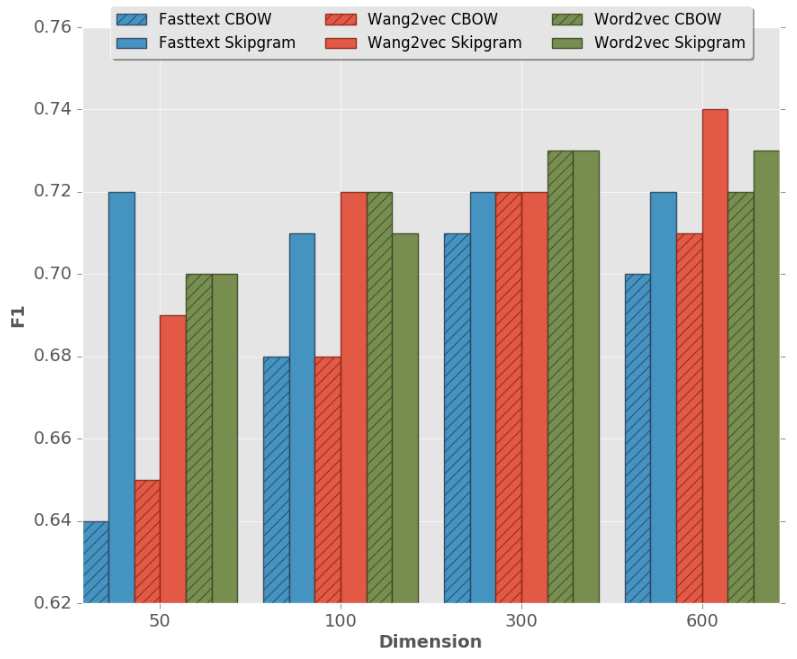} \label{fig:embs_ccl}}
\\
  \subfigure[AD]{\includegraphics[width=0.485\textwidth, keepaspectratio]{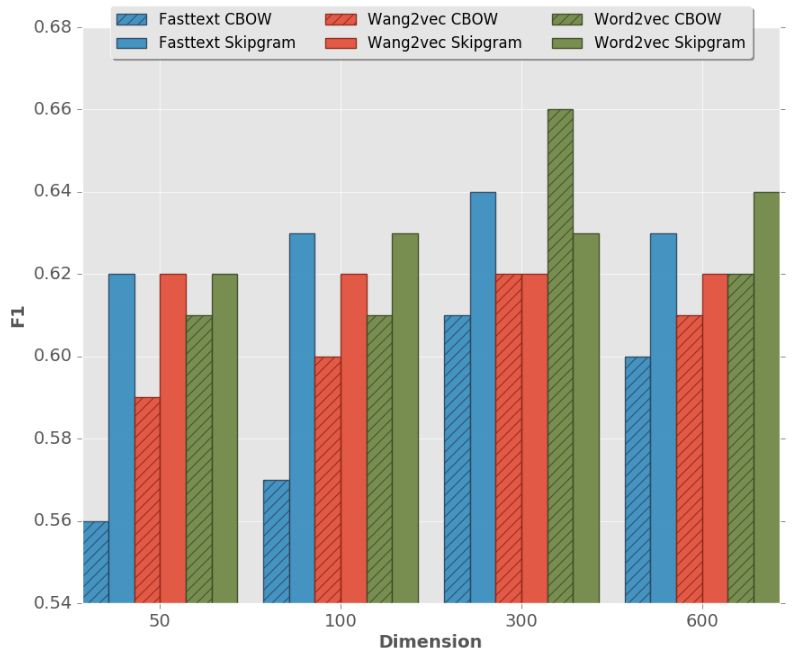} \label{fig:embs_da}}
  \subfigure[Constitution]{\includegraphics[width=0.485\textwidth, keepaspectratio]{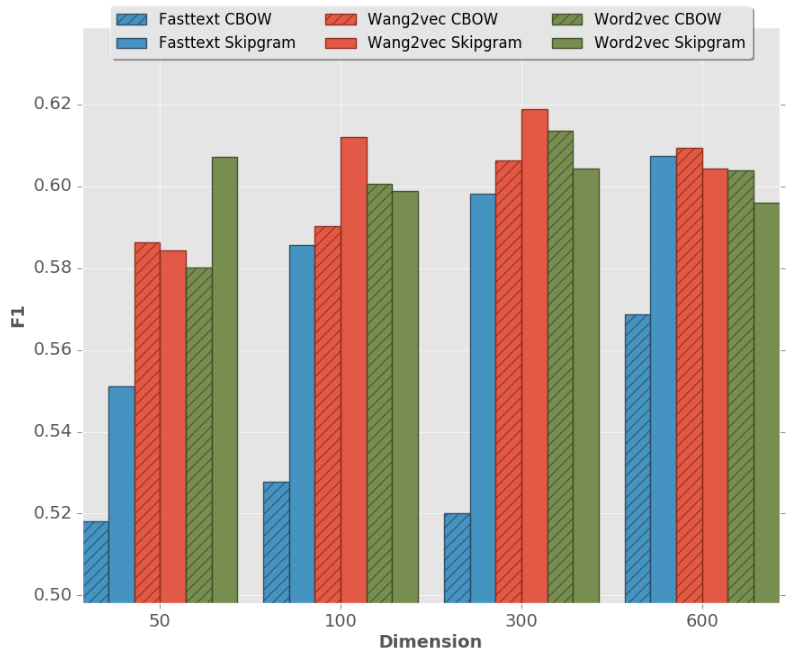} \label{fig:embs_const}}
  
  \caption{Results for different embedding methods and dimensions}
  \label{fig:embeddings_dimensoes}
\end{figure}

The performance results of the RCNN in terms of $F_1$ on each type of patient and on the Constitution dataset are shown in Figure \ref{fig:embeddings_dimensoes}, for which we vary the embedding methods and its training strategies along with the induced vector dimensions between the values of: $d \in \{50, 100, 300, 600\}$.

In most cases Word2vec achieved better performance than other methods. Specifically, for CTL with Skip-gram  and FastText with CBOW, yielding an $F_1$ of $0.76$, On the other hand, we see that for MCI patients, Wang2vec with Skip-gram was the best technique, yielding an $F_1$ of $0.74$. For the AD subjects the best technique was Word2vec with CBOW, returning an $F_1$ of $0.66$. As expected, results for CTL were higher than for MCI and AD, since the CTL narratives contain less noise. For Constitution data our method performs better using Wang2vec with Skip-gram strategy: $F_1$ of $0.62$.

It is possible to see in Figure \ref{fig:embeddings_dimensoes} that our method tends to better its performance with increasing dimension size. Furthermore, the Skip-gram strategy generally returned better results than CBOW for the FastText and Wang2vec methods, whereas for Word2Vec there were some variations when strategies were switched. Still, the Word2vec Skip-gram with 600 dimensions and CBOW with 300 dimensions were those which returned the best results for spontaneous and/or impaired speech (CTL, MCI and AD). In the case of the Constitution dataset, which is characterized as prepared and read speech, the best results were achieved by Wang2vec Skip-gram with 300 dimensions.

Contrary to the results reported in \cite{treviso2017sentence} using textual and prosodic information, our method obtained better performance for impaired speech transcriptions than for prepared speech. This is probably due to the fact that the Constitution includes more impacting prosodic clues, whereas for spontaneous/impaired speech, the lexical clues are of greater influence for classification. This difference between lexical and prosodic features for prepared and spontaneous speech is consistent with the finding reported in other studies \cite{Kolar2009,Fraser2015,treviso2017sentence}.

\section{Conclusion and Future Work}

Our objective in this work was to identify the embedding with the best performance for SBD, specifically whether it would be one which captures semantic information, like Word2vec; syntactic, like Wang2vec; or morphological, like FastText. Still, we were not able to discern which type was most influential in general, since the differences from one to another are very small. Also even when one technique was superior to another for a particular set, we still need to investigate whether this was actually the fault of the technique or due to secondary factors, like hyperparameters, random initialization, or even the conditions of the data used. 

In general, our results show that using only embeddings the RCNN method achieved similar results (difference of 1\%) to the state of the art in terms of $F_1$, using the same method published in \cite{treviso2017sentence} for both classes: CTL and MCI using embeddings with $600$ dimensions and prosodic information\footnote{Results using embeddings with 600 dimensions were obtained from the authors of the original article}. However, the results for the Constitution dataset were considerably lower (a difference of 17\%) than the results of the model which uses both lexical and prosodic information in conjunction, but the difference is less (4\%) for the models which used only prosodic information. Summing up, this indicates that by using a good word embedding model to represent textual information it is possible to achieve similar results with the state-of-the-art for impaired speech. 

Future work will include some investigation of the lexical and prosodic clues which impact the classification. Also, we would like to investigate whether disfluency detection in conjunction with SBD can yield better results.
Since the method presented in this paper can easily be applied to any language, we plan to evaluate it using English language corpora in order to directly compare the results with the related work.


\bibliographystyle{sbc}
\bibliography{biblio}

\end{document}